%% file: sample-sigconf.tex
\documentclass[sigconf]{acmart}

\usepackage{booktabs} 
\usepackage[ruled,vlined]{algorithm2e}
\usepackage{subcaption}

\begin{document}
\title{Population-Based Evolution Optimizes a Meta-Learning Objective}

\author{Kevin Frans}
\affiliation{%
  \institution{Cross Labs, Cross Compass Ltd.}
  \country{Japan}
}
\affiliation{%
  \institution{Massachusetts Institute of Technology}
  \city{Cambridge}
  \state{MA}
  \country{USA}
}
\author{Olaf Witkowski}
\affiliation{%
  \institution{Cross Labs, Cross Compass Ltd.}
  \country{Japan}
}
\affiliation{%
  \institution{Earth-Life Science Institute, Tokyo Institute of Technology}
  \country{Japan}
}
\affiliation{%
  \institution{College of Arts and Sciences, University of Tokyo}
  \country{Japan}
}


\begin{abstract}
Meta-learning models, or models that learn to learn, have been a long-desired target for their ability to quickly solve new tasks. Traditional meta-learning methods can require expensive inner and outer loops, thus there is demand for algorithms that discover strong learners without explicitly searching for them. We draw parallels to the study of evolvable genomes in evolutionary systems -- genomes with a strong capacity to adapt -- and propose that meta-learning and adaptive evolvability optimize for the same objective: high performance after a set of learning iterations. We argue that population-based evolutionary systems with non-static fitness landscapes naturally bias towards high-evolvability genomes, and therefore optimize for populations with strong learning ability. We demonstrate this claim with a simple evolutionary algorithm, Population-Based Meta Learning (PBML), that consistently discovers genomes which display higher rates of improvement over generations, and can rapidly adapt to solve sparse fitness and robotic control tasks.
\end{abstract}

%
%


\maketitle

\input{samplebody-conf}

\bibliographystyle{ACM-Reference-Format}
\bibliography{sample-bibliography} 

\end{document}

%% file: samplebody-conf.tex
\section{Introduction}

In recent years, there has been a growing interest in the community about meta-learning, or learning to learn. In a meta-learning viewpoint, models are judged not just on their strength, but on their capacity to increase this strength through future learning. Strong meta-learning models are therefore desired for many practical applications, such as sim-to-real robotics \cite{sim2real, sim2real2, sim2real3} or low-data learning \cite{maml, lowdata, lowdata2, lowdata3}, for their ability to quickly generalize to new tasks. The inherent challenge in traditional meta-learning lies in its expensive inner loop -- if learning is expensive, then learning to learn is orders of magnitude harder. Following this problem, there is a desire for alternate meta-learning methods that can discover fast learners without explicitly searching for them.

In the adjacent field of evolutionary computation, a similar interest has risen for genomes with high evolvability. Genomes represent individual units in a world, such as the structure or behavior or a creature, which together form a global population. An evolutionary system improves by mutating this population of genomes to form offspring, which are then selected based on their fitness scores. A genome’s evolvability is its ability to efficiently carry out this mutation-selection process \cite{ebility1, ebility2}. Precise definitions of evolvability are debated: some measurements capture the diversity of a genome’s offspring \cite{diversity1, diversity2, biology}, while others capture the ability for offspring to adapt to new challenges \cite{inev3, inev2, dynamic}.

From a meta-learning perspective, the ability for offspring to adapt to new challenges -- which we refer to as \emph{adaptive evolvability} -- is of particular interest. Consider a shifting environment in which genomes must repeatedly adapt to maintain high fitness levels. A genome with high adaptive evolvability would more easily produce offspring that explore towards high-fitness areas. Thus, adaptive evolvability can be measured as the expected fitness of a genome's descendants after some number of generations.

\begin{figure}
    \centering
    \includegraphics[width=\linewidth]{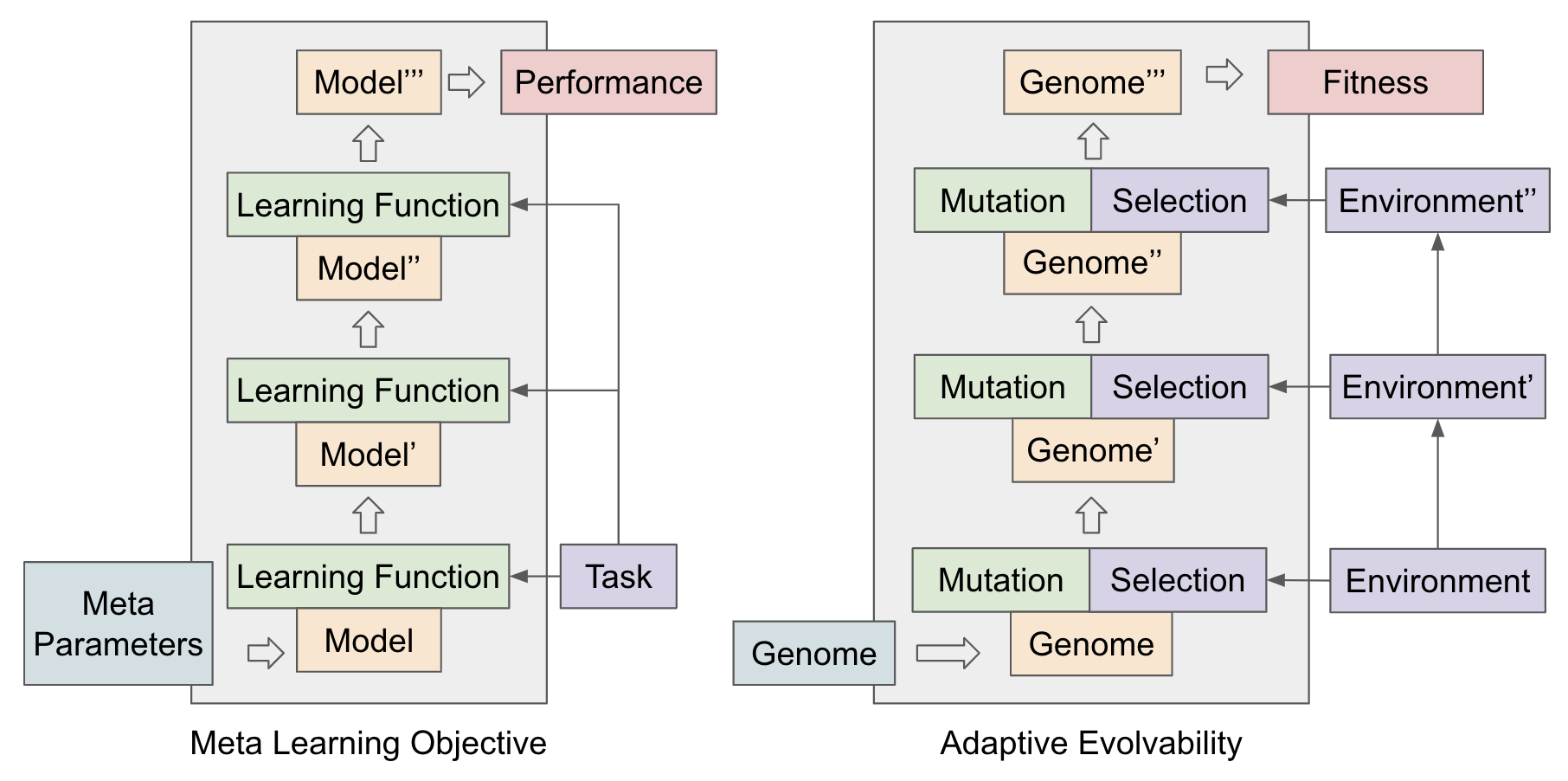}
    \caption{Meta-learning and adaptive evolvability both measure parallel objectives: performance after a set of learning iterations. An evolutionary system can be considered as "learning to learn" if its population of genomes consistently improve their evolvability, thus allowing the population to more efficiently adapt to changes in the environment.}
    \label{fig:comparison}
\end{figure}

Here we consider the evolutionary system from an optimization perspective. Evolutionary systems are learning models which iteratively improve through mutation and selection. Thus, the learning ability of a system directly depends on its genomes’ ability to efficiently explore via mutation. Take two populations and allow them to adapt for a number of generations; the population with higher learning ability will produce descendants with higher average fitness. The meta-learning objective of an evolutionary system is therefore the same as maximizing the adaptive evolvability of its population -- genomes should learn to produce high-fitness descendants through mutation (Figure \ref{fig:comparison}).

The key insight in connecting these ideas is that evolvability is known to improve naturally in certain evolutionary systems. In contrast to meta-learning, where meta-parameters are traditionally optimized for in a separate loop \cite{metasurvey}, evolvability is often a property that is thought to arise indirectly as an evolutionary process runs \cite{inev1, inev2, inev3, inev4}, a key example being the adaptability of life on Earth \cite{biology}. A large body of work has examined what properties lead to increasing evolvability in evolutionary systems \cite{inev1, inev2, inev3, inev4, inev5}, and if a population consistently increases in evolvability, it is equivalent to learning to learn.

In this work, we claim that in evolutionary algorithms with a large population and a non-static fitness landscape, high-evolvability genomes will naturally be selected for, creating a viable meta-learning method that does not rely on explicit search trees. The key idea is that survivable genes in population-based evolutionary systems must grant fitness not only in the present, but also to inheritors of the gene. In environments with non-static fitness landscapes, this manifests in the form of learning an efficient mutation function, often with inductive biases that balance exploration and exploitation along various dimensions. 

We solidify this claim by considering Population-Based Meta Learning (PBML), an evolutionary algorithm that successfully discovers genomes with a strong capacity to improve themselves over generations. We show that learning rates consistently increase even when a genome’s learning behavior is decoupled from its fitness, and genomes can discover non-trivial mutation behavior to efficiently explore in sparse environments. We show that genomes discovered through population-based evolution consistently outperform competitors on both numeric and robotic control tasks.

\section{Background}
\subsection{Meta Learning}

The field of meta-learning focuses on discovering models that display a strong capacity for additional learning, hence the description learning to learn \cite{learn2learn}. Traditionally, meta-learning consists of a distribution of tasks, along with an inner and outer loop. In the inner loop, a new task is selected, and a model is trained to solve the task. In the outer loop, meta-parameters are trained to increase expected performance in the inner loop. A strong meta-learning model will encode shared information in these meta-parameters, so as to quickly learn to solve a new task.

Meta-parameters can take many forms \cite{metasurvey}, as long as they affect inner loop training in some way. Hyperparameter search \citep{hyperparam0, hyperparam1, hyperparam2} is a simple form of meta-learning, in which parameters such as learning rate are adjusted to increase training efficiency. Methods such as MAML \cite{maml, maml2, maml3} and Reptile \citep{reptile} define meta-parameters as the initial parameters of an inner training loop, with the objective to discover robust representations that can be easily adapted to new problems. Others view meta-parameters as specifications for a neural architecture, a subset known as Neural Architecture Search \cite{arch1, arch2, arch3}. Still other meta-learning algorithms optimize the inner learning function itself, such as defining the learning function as a neural network \cite{rl2, rl3, rl4}, or adjusting an auxiliary reward function in reinforcement learning \cite{loss1, loss2, loss3}. In our method, meta-parameters are included in the genome, which both defines a starting point for mutation and can also parametrize the mutation function itself.

A common bottleneck in meta-learning is in the computational needs of a two-loop process. If the inner loop requires many iterations of training, then the outer loop can become prohibitively expensive. Thus, many meta-learning algorithms focus specifically on the one-shot or few-shot domain to ensure a small inner loop \cite{lowdata, lowdata2, lowdata3}. On many domains, however, it is unrealistic to successfully solve a new task in only a few updates. In addition, meta-parameters that are strong in the short term can degrade in the long term \cite{longterm}. In this case, other methods must be used to meta-learn over the long term, such as implicit differentiation \cite{implicit}. The approach we adopt is to combine the inner and outer loops, and optimize parameters and meta-parameters simultaneously, which can significantly reduce computation cost \cite{onlinelr, simul1, simul2, simul3}. The key issue here is that models may overfit on their current tasks instead of developing long-term learning abilities. We claim that evolving large populations of genomes reduces this problem, as shown in experiments below.

\subsection{Evolvability}
The definition of evolvability has often shifted around, and while there is no strict consensus, measurements of evolvability generally fall into two camps: diversity and adaptation. Diversity-based evolvability focuses on the ability for a genome to produce phenotypically diverse offspring \cite{diversity1, diversity2, biology, ebility1}. This measurement, however, often depends on a hand-specified diversity function. In the context of learning to learn, diverse offspring are not always optimal, rather diversity is only a factor in an exploration-exploitation tradeoff to produce high-fitness offspring \cite{exploreexploit}. Instead, we consider adaptation-based evolvability \cite{inev3, inev2, ebility2}, which measures the performance of offspring some generations after a change in the fitness landscape.

As evolvability is a highly desired property in evolutionary algorithms, many works aim to explain how evolvability emerges in evolutionary systems. Some works attribute evolvability due the rules of the environment. Periodically varying goals have been shown to induce evolvability in a population of genomes, as genomes with more evolvable representations can adapt quicker to survive selection \cite{dynamic, inev3}. In addition, environments where novelty is rewarded or where competition is spread around niches will naturally favor genomes that produce diverse offspring \cite{inev4}. Others take a different viewpoint, and instead consider the genetic algorithm and representations in use as a major cause of evolvability. Specifically, research into evolvable representations has shown that there are benefits to modularity and redundancy in genomes \cite{ebility2}. Exploring neutral landscapes, where multiple genotypes map to the same phenotype, has also been shown to aid in mutation diversity \cite{inev2}. A few works even take a step further and claim that given certain conditions, the appearance of evolvability is inevitable \cite{inev4, inev5}, a view we further examine in this work. 

\subsection{Evolutionary Meta Learning}
Many methods take inspiration from both evolutionary systems and meta-learning, in most cases using evolution as a gradient-less optimizer for either the inner or outer meta-learning loop. Evolution in the outer loop has been shown to successfully learn policy gradient algorithms \citep{evolvepg}, loss functions \cite{evolveloss}, neural network structures \cite{evolvestruct}, and initial training parameters \cite{evolvemaml}. There has been little focus, however, on viewing evolutionary systems as both the inner and outer loop in a meta-learning formulation, where a genome evolves \emph{while} improve its own learning ability.

In addition, a recent trend is to use the method known as Evolution Strategies \cite{evolveopenai}, which involves evolving a global genome by mutating a population of offspring, then updating the genome based on a weighted average of the offspring's parameters. This method has proven the scalability of evolutionary methods, however, it remains a greedy algorithm, which as shown in experiments below fail to select for long-term benefits.

\section{Approach}
Our motivation is to merge the benefits of evolutionary algorithms and meta-learning into a single method. Genome that display strong adaptive abilities -- genomes that have learned to learn -- are heavily desired in areas such as open-endedness \cite{openend, openend2} and quality diversity \cite{qualitydiversity}, as they can efficiently explore new niches and ideas. Traditional meta-learning methods, however, are often bottlenecked by expensive inner and outer loops, and optimization can be limited by gradient degradation. Rather, we would prefer to continuously train a model on new tasks, and have the model naturally improve its learning ability over time.

We present Population-Based Meta Learning (PBML), an easily specified evolutionary algorithm that favors genomes that quickly adapt to changes in the fitness landscape.

\begin{algorithm}
\SetAlgoLined
Define an initial set of genomes G. Each genome contains parameters, as well as a population count.\
Define a fitness function F(), which evaluates the fitness of a genome.\;
Define a stochastic mutation function M(), which creates offspring by mutating the parameters of a genome.\;
Define decay ratio D; the ratio of population that dies every generation.\;
Define offspring count C, the total number of offspring produced every generation.\;
Define ranking function Rank(), which outputs a genome's floating-point rank between 0 and 1 based on fitness.\;
\For{timestep=0...N}{
    \For{g in G} {
        g.fitness = F(g.parameters)\;
        g.rank = Rank(g.fitness)\;
        g.population *= (1-D)\;
        g.population *= g.rank\;
        g.population /= total\_population\;
        \For{child in int(g.population * C)} {
            G.append(M(g))\;
        }
    }
}
\caption{Population-Based Meta Learning}
\end{algorithm}

\subsection{Reasoning}

We first define evolutionary systems as methods that iteratively select for genomes with high fitness. In a standard evolutionary system, we consider a population of genomes along with a fitness function. Every generation, genomes that display high fitness increase in population size, while genomes with lower fitness die off. Exploration is introduced in the form of descendants -- each genome occasionally mutates, producing offspring with slight variations. Over generations, the population will skew towards genomes that display high fitness.

We now address the meta-learning objective of such a system. The traditional meta-learning objective is to optimize for performance after a series of learning iterations. In evolutionary systems, this learning iteration is a combination of mutation and selection. A simple visualization of this is a single genome which mutates to produce 100 offspring, where the child with the highest fitness survives. This can be seen as a one-step lookahead search towards the peak of a fitness landscape. As such, the meta-learning performance of a genome can be defined as the fitness of its offspring after a number of generations. This is the same definition as adaptation-based evolvability, a parallel which guides us moving forwards.

A genome that has “learned to learn” must in fact “learn to mutate”. The simplest form of learning to mutate involves developing a robust genetic representation. Consider an environment where the mutation function of a genome consists of noise added to its parameters. In this case, a genome where slight changes in parameters result in large changes in the displayed phenotype would be able to explore a larger space than its counterparts. The mutation function may also be directly parametrized in the genome, such as a parameter defining mutation radius. Crucially, producing diverse offspring through mutation is not always optimal; there is an exploration-exploitation tradeoff. However, some dimensions may consistently be worth exploring -- e.g. variation in dental structure helps adapt to new food types, but variation in blood proteins is often lethal. A strong meta-learned genome would encode these inductive biases into their mutation function, and explore in dimensions that are useful for improving fitness.

The key intuition behind PBML is that evolutionary systems with large populations naturally optimize for long term fitness. As a simple example, consider two genomes, Red and Blue. Red has higher fitness than Blue, but it also has a gene that causes 50\% of its offspring to mutate lethally. Over time, descendants of Red will die at higher rates, so the descendants of Blue will become more present in the population. In other words, a gene with high survivability not only grants high fitness to its genome, but also must maintain this fitness advantage in the offspring that inherit it. Thus, over generations, the genes that survive will be genes that allow their inheritors to achieve high fitness -- precisely the definition of a gene with strong meta-learning capabilities.

An important requirement for meta-learning to be selected for is that a variety of genomes must be allowed to survive each generation. In evolutionary systems, genomes compete with each other to survive by increasing their fitness over generations. It is important that genomes with lower fitness are not immediately removed, so that competition for long-term fitness can emerge. Imagine a greedy evolutionary system where only a single high-fitness genome survives through each generation. Even if that genome’s mutation function had a high lethality rate, it would still remain. In comparison, in an evolutionary system where multiple lineages can survive, a genome with lower fitness but stronger learning ability can survive long enough for benefits to show. 

Large population numbers are helpful for two reasons: to counteract noise, and to capture selection effects. A genome with a strong mutation function will produce fitter offspring on average, but these effects are confounded with noise inherent to mutation. The larger the amount of offspring, the more accurately an evolutionary system will skew towards strong mutation functions. In addition, large population counts allow for finer-grained effects to be represented. For example, imagine the population of genome A is to fall by 7\%, and genome B to fall by 10\%. If only 10 copies of each genome are present, both genomes will fall to 9 population, and the difference will not be seen. In contrast, if 10,000 copies are present, then the population counts can be more smoothly adjusted.

In PBML, we approximate this behavior by defining each genome’s population as a ratio. This provides a continuous way to measure selection pressure due to fitness. A genome with lower fitness will decay its population ratio faster than a higher fitness genome. However, they will remain in the simulation until they drop below a certain cutoff percentage. As genomes with higher populations will produce more offspring, the overall population will gradually skew towards high-fitness genomes, while allowing many lineages of genomes to be compared across generations.

\section{Experiments}

In a series of experiments, we examine how population-based evolution can discover genomes with strong learning ability. In the Numeric Fitness world, we show that population-based methods optimize for long-term improvement, even if current fitness is decoupled from improvement ability. In the Square Fitness world, we show that genomes can successfully learn mutation functions which efficiently explore areas with high potential fitness, while avoiding zero-fitness areas. Finally, we display that population-based evolution discovers robotic control policies that are easily adaptable to new goals.

\subsection{Do population-based evolutionary systems select for genomes with long-term fitness improvement?}

\begin{figure}
    \includegraphics[width=0.9\linewidth]{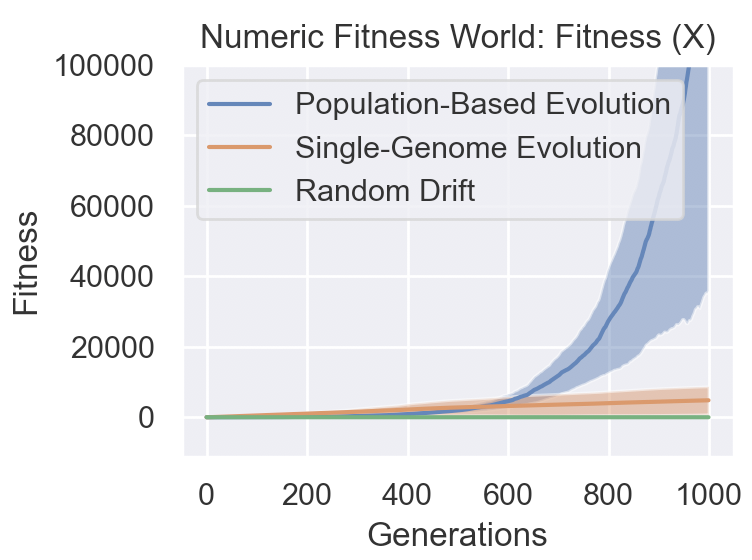}
     \includegraphics[width=0.9\linewidth]{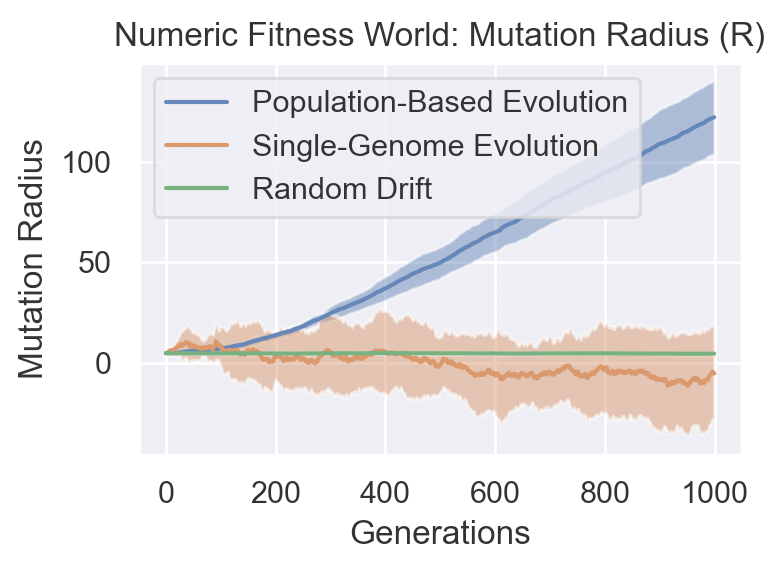}
    \caption{Numeric Fitness world, where current fitness (X) is decoupled from long-term learning ability (R). Population-based evolution results in genomes where learning ability steadily increases, as shown in a quadratic learning curve (top) and increasing mutation radius (bottom). In contrast, greedy single-genome evolution and random drift show no clear improvement in learning ability.}
    \label{fig:numeric}
\end{figure}

We first present an environment where a genome's long-term learning ability is decoupled from its current fitness, which we refer to as the Numeric Fitness world. We define a genome as a set of two numbers, X and R. The fitness of a genome is determined entirely by its X value, with higher X values resulting in higher fitness. Learning ability, on the other hand, is encoded by R, which parametrizes the amount of noise inserted in the mutation function.

\begin{equation}
Genome = \{X,R\}
\end{equation}
\begin{equation}
Fitness(X,R) = X
\end{equation}
\begin{equation}
   Mutate(X,R) = \begin{cases}
      X = X + Normal() * 1.05^R \\
      R = R + Normal()
    \end{cases}\, 
\end{equation}

In this Numeric Fitness experiment, our goal is to examine whether long-term learning ability appears when only current fitness is selected for. Specifically, long-term learning is represented by the R parameter of the genome, which specifies the mutation radius of X. Genomes with a larger R value will create a larger deviation in their offspring's X values, and are therefore more likely to produce offspring with higher fitness than their peers. It is important that while R has an affect on the offspring produced by a genome, it has no affect on fitness in the present. Thus, if R steadily increases through the generations, we can claim that there is consistent pressure to improve mutation ability even with no direct fitness gain.

In figure \ref{fig:numeric} (top), we show that population-based evolutionary systems steadily increase in their fitness improvement rate across generations. In the population-based system, 1000 offspring are produced each generation, with high-population genomes producing more offspring. Every generation, the population ratios of existing genomes drop from 25\% to 100\% depending on their fitness rank, and genomes with a ratio of less than 1/1000 are removed. In the greedy single-genome system, only the highest fitness genome survives through each generation, and all 1000 offspring are mutations of this single genome. The random drift system is the same as the population-based system, except all genomes are ranked as equal fitness. Notably, while in the beginning the single-genome system achieves higher average fitness, over time the genomes created through population-based evolution outperform the greedy variety.

Figure \ref{fig:numeric} (bottom) demonstrates the reasoning behind the learning ability gap. In a population-based evolutionary system, the survivability of a genome not only depends on current fitness but also on the fitness of its descendants. This results in a consistent pressure to improve learning ability through generations, which manifests in the form of a consistently increasing R parameter. In contrast, both greedy single-genome evolution and random drift display no significant trend in the R parameter, as a greedy system optimizes only for fitness within the current generation, and in a random drifting system there is no selection pressure at all.

\subsection{Can selection pressure for long-term fitness result in encoding useful inductive biases within a genome’s mutation function?}

\begin{figure}
  \includegraphics[width=\linewidth]{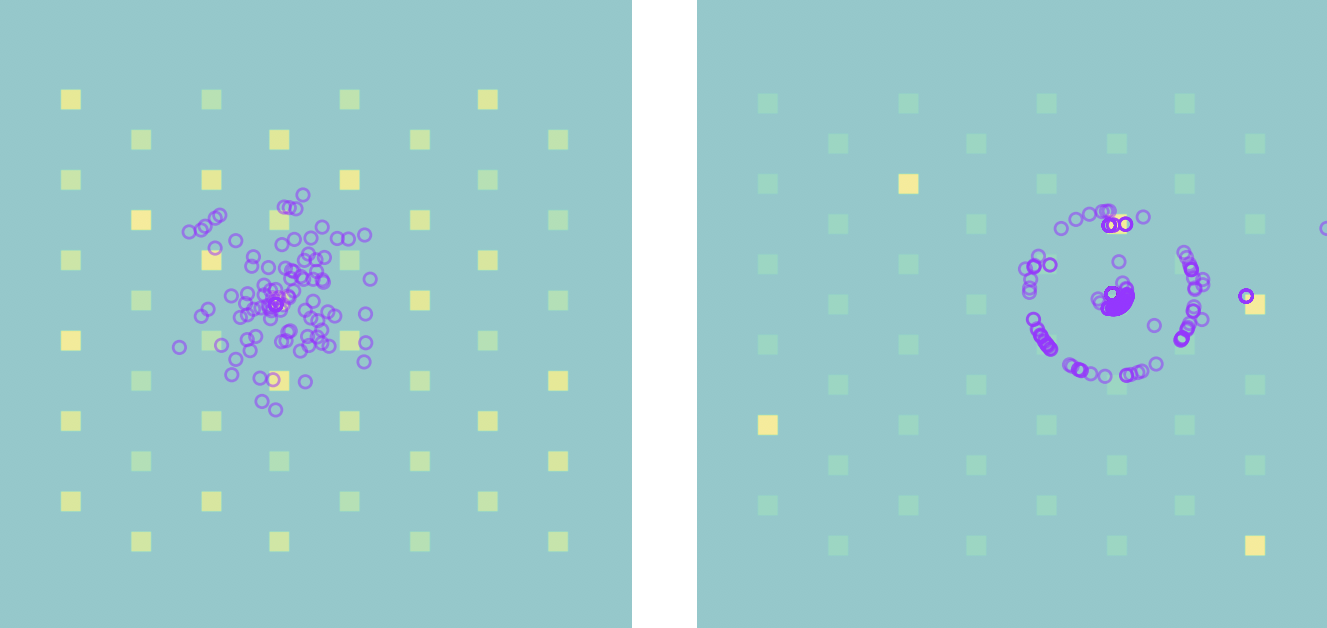}
\caption{
Left: Square Fitness world, where fitness-containing squares are uniformly spread throughout a gridworld. Every ten generations, the fitness value of each square is shuffled. Non-square areas have zero fitness. Right: Hard Square world, where squares only have a 10\% chance to be high-fitness.}
\label{fig:squarefitness}
\end{figure}

\begin{figure}
  \includegraphics[width=\linewidth]{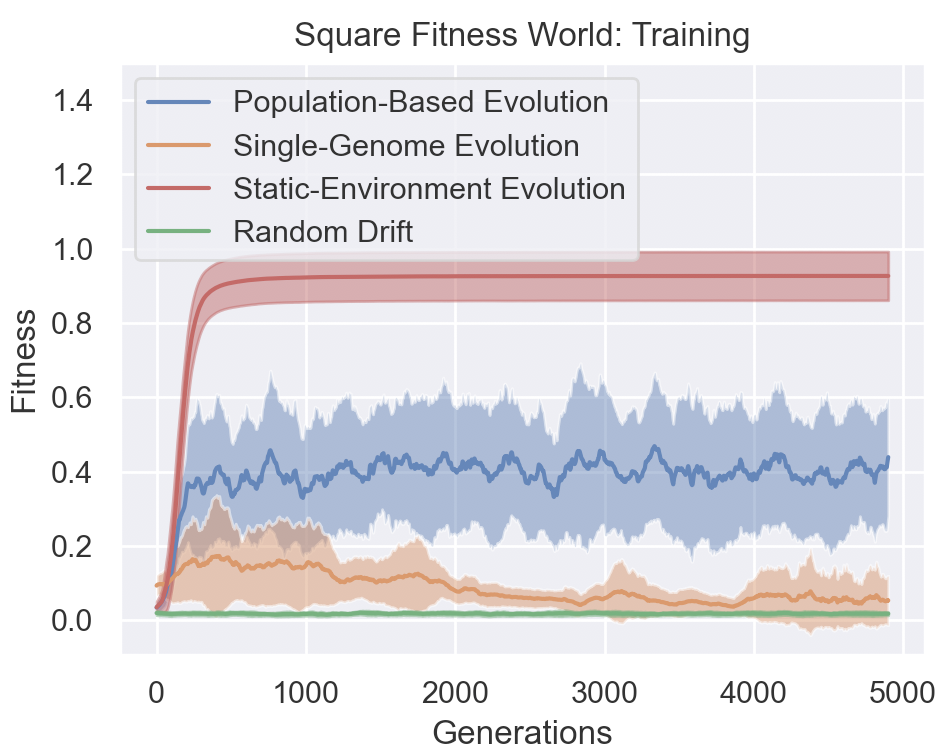}
  \includegraphics[width=\linewidth]{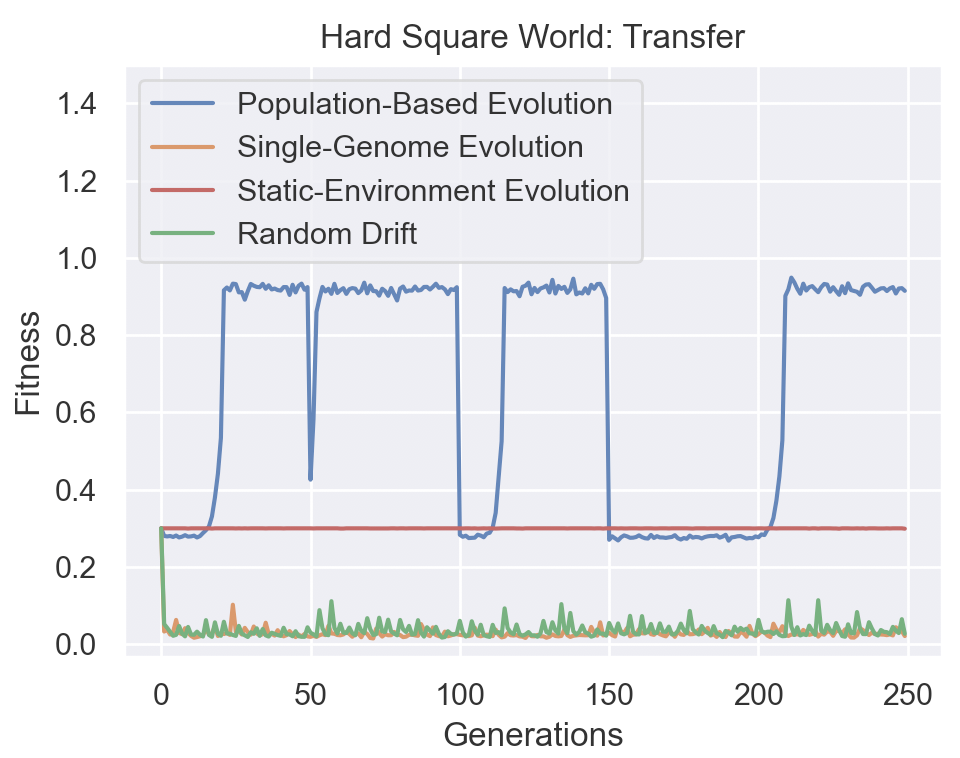}
\caption{
Genomes evolved in a Square Fitness world (top). After training, genomes are transferred to a Hard Square world (bottom). Population-based evolution results in genomes which develop efficient mutation functions, allowing them to consistently locate the high-fitness squares. Genomes evolved in a static environment instead learn to minimize mutation, and thus remain stuck in low-fitness areas when transferred. 
}
\label{fig:squarefitness}
\end{figure}


\begin{figure*}
  \includegraphics[width=.9\linewidth]{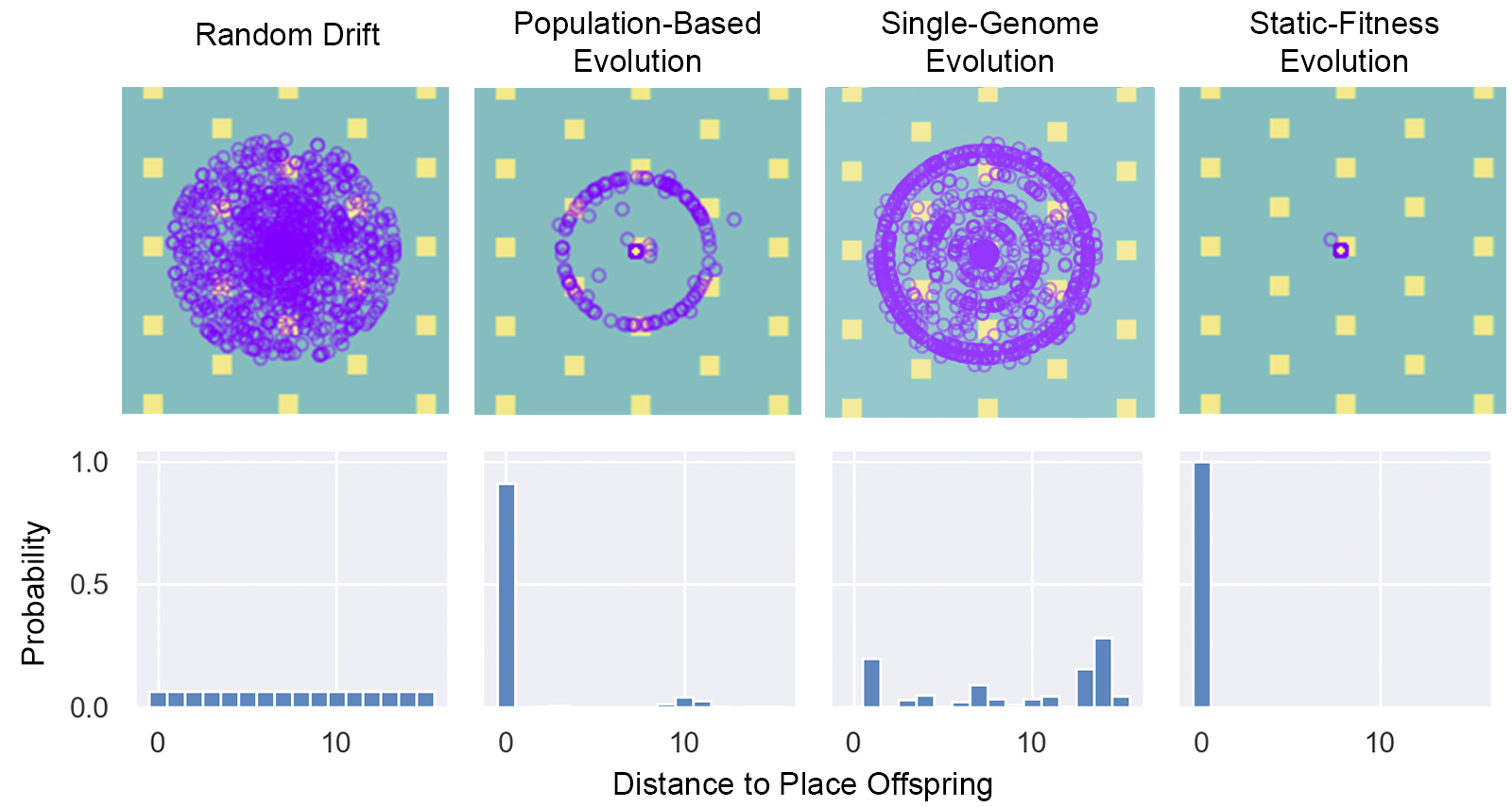}
\caption{Offspring produced by various genomes in Square Fitness world, according to their learned mutation behavior. 
Population-based evolution genomes produce offspring in a ring corresponding to the nearest set of neighboring squares, allowing for efficient exploration. Genomes in the static-fitness environment learn to reduce the effect of mutation, whereas mutation functions created from single-genome evolution and random drift show no clear behavior.}
\label{fig:mutations}
\end{figure*}

In a second experiment, the Square Fitness world, we consider a setup where a genome parametrizes its own mutation function in a non-trivial manner. A genome is defined by an X and Y coordinate along with a 16-length vector R that specifies its mutation behavior. Specifically, a genome mutates by creating a child a certain distance away from its XY coordinate, with the direction chosen randomly. When a genome mutates, each parameter in its R vector determines the probability that a child will be created a certain distance away.

\begin{equation}
Genome = \{X,Y,R1,R2,R3...R16\}
\end{equation}
\begin{equation}
    Mutate(X,Y,R) = \begin{cases}
        Radius = Softmax(R1,R2,R3...R16) \\
	    Direction = Random(360) \\
	    X += Cos(Radius) * Direction \\
	    Y += Sin(Radius) * Direction \\
    \end{cases}\, 
\end{equation}

The fitness landscape in this world is defined as a 256x256 coordinate grid, with a fitness value for every coordinate. Fitness is distributed as a series of spaced-out squares. Within each square, fitness can range between 0.3 to 1, whereas fitness remains at 0 in the areas in between. Importantly, a given square always has six neighbors within a fixed distance of itself. Every few generations, the fitness values of each square shuffle, while positions remain the same.

In this experiment, a successfully meta-learned genome would smartly define its mutation function to explore efficiently. It is important to be able to explore different squares to find the areas with the highest fitness. It is inefficient, however, to simply randomly explore around a coordinate, as most of the fitness landscape consists of zero-fitness area. Rather, a strong mutation function will only create offspring that land on a variety of other squares.

Figure \ref{fig:mutations} shows the mutation behaviors learned from various evolutionary methods. In the start, genomes have a uniform mutation function at every radius. After running population-based evolution, the genomes instead develop a prominent bias in their mutation function. Specifically, they place a large chunk of probability in maintaining their current coordinate, which can be seen as exploitation. They additionally, however, place some probability in producing offspring with a radius of 9-10 away from themselves -- which precisely matches the distance to a neighboring square.

Notably, this effect does not appear when dealing with certain variations. First, in a greedy single-genome evolutionary system, there is no pressure on adjusting the mutation function. As such, the general mutation behavior of the population drifts randomly. While we see rings, these rings are an artifact of the homogeneous population, and they do not correspond to any meaningful distance.

The second ablation compares to a population-based evolutionary system in which the fitness landscape is held static. In this case, we do in fact see pressure to improve long-term fitness, however, this pressure does not manifest in a mutation function that can be considered a strong learner. Instead, the population has learned to favor genomes which display as little diversity as possible. Intuitively, the genomes in this population have already achieved the optimal phenotype, and learn to continuously exploit this rather than exploring further.

\subsection{Do strong mutation functions enable faster learning on new, unseen tasks?}

A common goal in meta-learning is to create models that can quickly solve new tasks. In general, models are first trained on a distribution of related tasks so that general meta-knowledge can be extracted. This meta-knowledge is then used to efficiently learn the specifics of a new task.

We utilize this framework to test the learning capabilities of genomes evolved in the Square Fitness environment. We define a new environment, referred to as the Hard Square environment, in which each square has only a 10\% chance to contain high (1.0) fitness. All other squares instead contain low (0.3) fitness, and all non-square areas contain zero fitness. Additionally, high-fitness squares are limited to the outer areas; thus squares near the center are always low fitness. This task poses a challenging exploration problem, as high-fitness areas are sparsely located, and there is no gradient to inform a direction of improvement.

Figure \ref{fig:squarefitness} shows that population-based evolution genomes significantly outperform others when transferred. This is likely due to the biases encoded in their mutation function -- by exploring only in the spaces where squares are present, they are much more likely to discover high-fitness areas. In contrast, genomes evolved in a static-fitness environment demonstrate their weakness -- they have learned to reduce exploration, and thus fail to escape the local minimum of the central low-fitness squares. Genomes evolved in single-genome evolution and random drift fail to achieve even this baseline, as their inefficient mutation functions result in a large percentage of offspring being born into zero-fitness areas.

\subsection{Can population-based evolution help to efficiently adapt robotic control policies?}

Finally, we wish to examine if population-based evolution can meta-learn in a domain with more complex dynamics. We consider the Reacher-v2 environment, as defined in OpenAI gym \cite{openaigym}, where a Mujoco-simulated \cite{mujoco} arm must maneuver its tip to a point on a grid. The arm is controlled by adjusting the torque of two joints, and information about the current position and angles of the arm is given as an observation every frame. Fitness is calculated as the distance between the arm and the goal point over 50 frames. Every fifty generations, a new goal point is sampled, forcing genomes to quickly adapt their policies to the new goal.

To allow for genomes to develop meta-learning capabilities, each genome is defined as a four-layer neural network. The second and third layers are comprised of three independent feedforward modules, which are summed according to a parametrized softmax function to form the full output of the layer. Every module has an independent mutation radius, encoded as a 12-parameter vector stored in the genome. 

We define the genome this way to enable two avenues for improving learning: robust representations and smart mutation functions. Neural networks map a set of parameters to a policy. There are multiple networks that can map to the same policy, however some networks may be better at adapting than others. To increase learning ability, a genome may learn a robust representation that can easily mutate to produce diversity. In addition, a genome can adjust the mutation radiuses of its modules to develop a fine-tuned exploration behavior, such as lowering mutation in a module that parses observations, but increasing mutation in a module that calculates where to position the arm.

\begin{figure}
  \includegraphics[width=\linewidth]{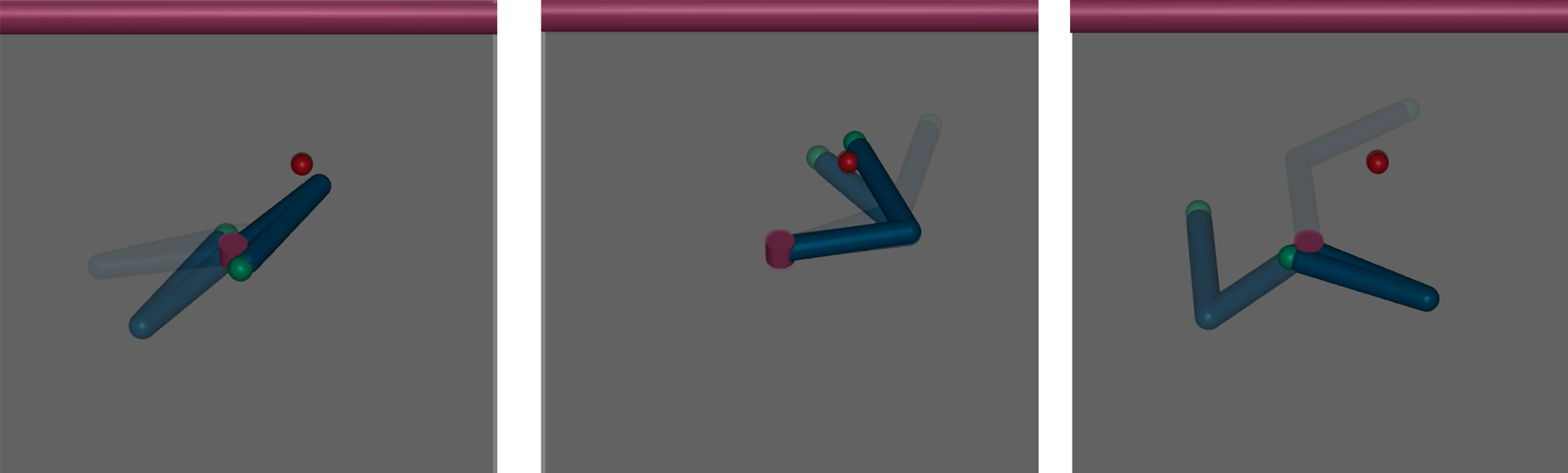}
\caption{Robotic control task, in which a two-joint arm must be navigated to a goal point. In all methods except the static-environment system, the goal location is randomized every 50 generations. Genomes define a neural network policy which is given observations about the arm's position and velocity, and controls the force applied to each joint.}
\label{fig:mujoco_policy}
\end{figure}

\begin{figure}
  \includegraphics[width=\linewidth]{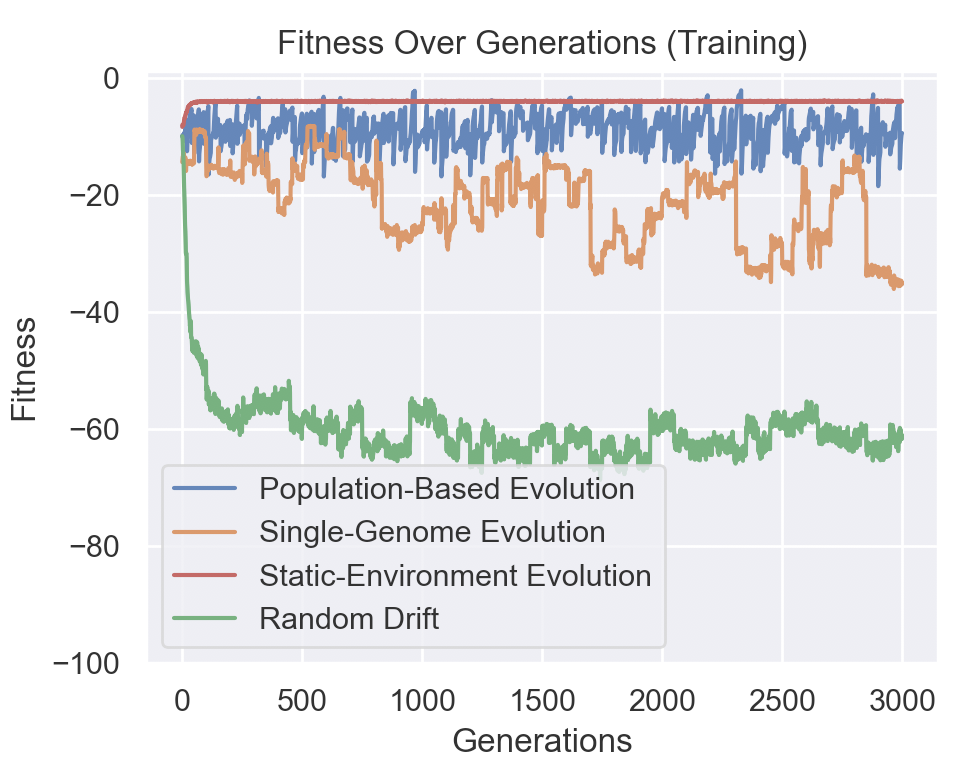}
\caption{Training on the robotic control task. Population-based evolution manages to discover genomes which can adapt to new goal points; whereas single-genome evolution and random drift genomes develop unstable mutation functions that hinder learning.}
\label{fig:mujoco_train}
\end{figure}

Figure \ref{fig:mujoco_train} shows that population-based evolution generally maintains a steady adaptation rate during training, while single-genome evolution results in more unstable behavior. To further measure the learning ability of the various genomes, we then transfer each genome to an unseen test goal. Table \ref{tab:mujoco_test} showcases the advantage and drawbacks of the population-based systems. Notably, the population-based and static-environment genomes learn to constrain their mutation functions, and can maintain strong policies through many generations. In contrast, offspring from genomes in the other methods have a high chance to mutate badly, lowering their average fitness. This constraint, however, comes at a slight cost. The population-based genome comes close to the top-performing policy but falls short, as it likely has stopped mutation in a module that is slightly sub-optimal. This can be seen as a exploration-exploitation tradeoff, in which the genome gains a higher average fitness by constraining its search space of offspring, but can fail when the optimal solution is outside the space it considers.

\begin{table}
  \caption{Fitness after 50 generations on a new goal}
  \label{tab:mujoco_test}
  \begin{tabular}{ccl}
    \toprule
    Method&Top Fitness&Average Fitness\\
    \midrule
    Population-Based Evolution & -2.962 & -2.966\\
    Single-Genome Evolution & -10.835 & -34.591\\
    Static-Environment Evolution & -5.197 & -5.199\\
    Random Drift & -3.585 & -8.829\\
    From Scratch & -2.2415 & -9.982\\
  \bottomrule
\end{tabular}
\end{table}

\section{Discussion}
In this work, we show that population-based evolutionary systems naturally optimize for genomes that display long-term learning ability. From a meta-learning perspective, this work sheds light on a new perspective for developing meta-learning algorithms that learn through constant peer competition rather than explicit inner and outer loops. This paper also contributes to the study of emergent evolvability. Works from the artificial life community have discussed how evolvability emerges through simulated systems \cite{inev4}, while works in biological systems have questioned why the DNA-based genomes of life on Earth display such high adaptive capacity \cite{biology}. The ideas we present create a solid perspective on how evolvability can emerge naturally through an evolutionary process over large populations of creatures. 

We believe this paper provides a starting point to many future pathways examining evolutionary systems as a form of meta-learning. A promising direction is to examine which populations retain their evolvability as evolution continues. As seen in experiments where the fitness landscape remains static, in certain environments a genome will lose its learning ability in return for a more consistent set of offspring \cite{inev5}. However, there may be situations where this does not occur -- for example, if a genome derives its evolvability by developing a robust representation, there would be no pressure to lose such a representation. We believe the question of “what genomes retain evolvability” is a promising future path.

Another direction of research lies in asking “what aspects of a population define its learning ability”. In this paper, we focused on genomes which learn strong mutation functions, so their offspring will have a higher expected fitness. Another avenue could be to instead learn a strong selection function. For example, rather than allowing only its high-fitness offspring to survive, a genome could encode an artificial selection function that rewards offspring with high diversity. In the long run, such a selection function could still result in higher fitness due to a higher capacity for exploration. 

We hope this paper serves as a conduit between the study of meta-learning and evolvability in evolutionary systems, which we believe to be heavily related. There is room for many future works in areas connecting these fields, and we believe that sharing ideas will result in further insight in both directions.


\appendix

\begin{acks}
Thanks to Lisa Soros for early feedback.

\end{acks}